\documentclass[10pt,twocolumn,letterpaper]{article}

\usepackage[pagenumbers]{cvpr}

\usepackage[T1]{fontenc}

\usepackage{booktabs}
\usepackage{multirow}
\usepackage{graphicx}
\usepackage{caption}
\usepackage{amsmath}
\usepackage{xspace}

\usepackage{listings}
\lstdefinelanguage{json}{
  basicstyle=\ttfamily\footnotesize,
  breaklines=true,
  breakatwhitespace=false,
  showstringspaces=false,
  literate=
    *{:}{{{\color{gray}:}}}{1}
     {,}{{{\color{gray},}}}{1}
     {\{}{{{\color{gray}\{}}}{1}
     {\}}{{{\color{gray}\}}}}{1}
     {[}{{{\color{gray}[}}}{1}
     {]}{{{\color{gray}]}}}{1}
}

\setlist{nosep,leftmargin=*}

\newcommand{\bench}{\textsc{Varex}\xspace}
\newcommand{\benchflat}{\textsc{Varex-Flat}\xspace}
\newcommand{\benchnested}{\textsc{Varex-Nested}\xspace}
\newcommand{\benchtable}{\textsc{Varex-Table}\xspace}

\newcommand{\mP}{\textbf{P}\xspace}
\newcommand{\mS}{\textbf{S}\xspace}
\newcommand{\mV}{\textbf{V}\xspace}
\newcommand{\mSV}{\textbf{S{+}V}\xspace}

\graphicspath{{figures/}}

\definecolor{cvprblue}{rgb}{0.21,0.49,0.74}
\usepackage[breaklinks,colorlinks,allcolors=cvprblue]{hyperref}

\def\paperID{}
\def\confName{CVPR}
\def\confYear{2026}

\title{VAREX: A Benchmark for Multi-Modal Structured Extraction from Documents}

\author{
Udi Barzelay\quad Ophir Azulai\quad Inbar Shapira\quad Idan Friedman\\
Foad Abo Dahood\quad Madison Lee\quad Abraham Daniels\\[2pt]
{\small IBM Research}\\[-1pt]
{\tt\scriptsize \{udib, ophir, inbar\_shapira, idan.friedman, foad.abo.dahood, madisonklee, abraham.daniels1\}@ibm.com}
}

\begin{document}
\maketitle
\begin{abstract}\vspace{-2pt}
We introduce VAREX (VARied-schema EXtraction), a benchmark for evaluating multimodal foundation models on structured data extraction from government forms. VAREX employs a Reverse Annotation pipeline that programmatically fills PDF templates with synthetic values, producing deterministic ground truth validated through three-phase quality assurance. The benchmark comprises 1,777 documents with 1,771 unique schemas across three structural categories, each provided in four input modalities: plain text, layout-preserving text (whitespace-aligned to approximate column positions), document image, or both text and image combined.
Unlike existing benchmarks that evaluate from a single input representation, VAREX provides four controlled modalities per document, enabling systematic ablation of how input format affects extraction accuracy, a capability absent from prior benchmarks. We evaluate 20 models\footnote{Two additional 2B models were evaluated but excluded from all tables due to near-zero extraction scores.} from frontier proprietary models to small open models, with particular attention to models $\leq$4B parameters suitable for cost-sensitive and latency-constrained deployment. Results reveal that (1)~below 4B parameters, structured output compliance (not extraction capability) is a dominant bottleneck; in particular, schema echo (models producing schema-conforming structure instead of extracted values) depresses scores by 45--65\,pp (percentage points) in affected models; (2)~extraction-specific fine-tuning at 2B yields $+$81\,pp gains, demonstrating that the instruction-following deficit is addressable without scale; (3)~layout-preserving text provides the largest accuracy gain ($+$3--18\,pp), exceeding pixel-level visual cues; and (4)~the benchmark most effectively discriminates models in the 60--95\% accuracy band.
Dataset and evaluation code are publicly available.\footnote{\url{https://huggingface.co/datasets/ibm-research/VAREX} \quad \url{https://github.com/udibarzi/varex-bench}}
\end{abstract}

\section{Introduction}
\label{sec:intro}

\begin{figure}[t]
  \centering
  \includegraphics[width=\columnwidth]{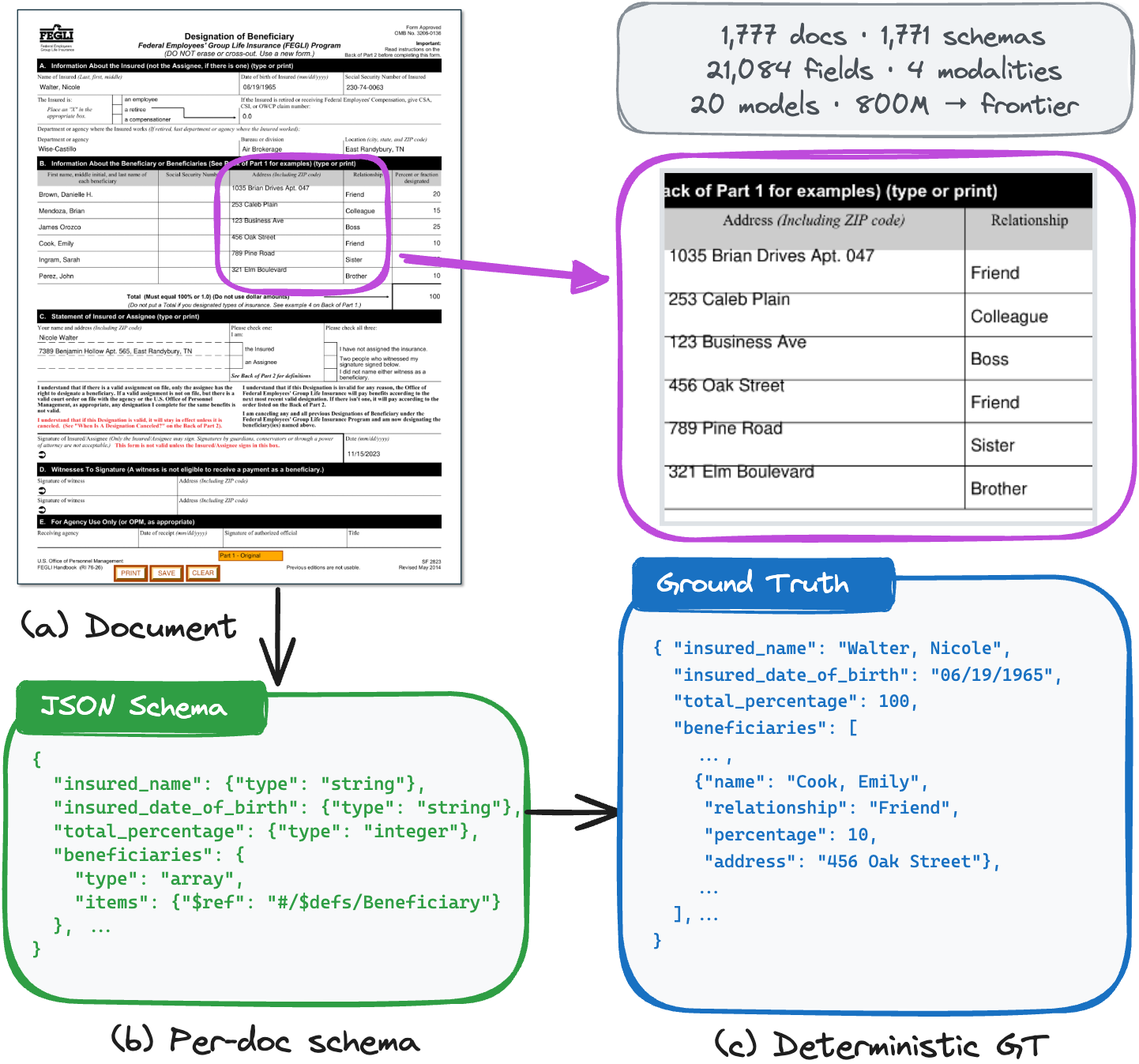}
  \caption{\bench benchmark overview. A government form~(a) is paired with a per-document JSON schema~(b) that defines the extraction target, including nested structures via \texttt{\$ref}. Forms are programmatically filled with realistic data, and ground truth~(c) is derived directly from the fill values. The benchmark spans 1,777 documents with 1,771 unique schemas.}
  \label{fig:teaser}
\end{figure}

\begin{figure*}[t!]
  \centering
  \includegraphics[width=\textwidth]{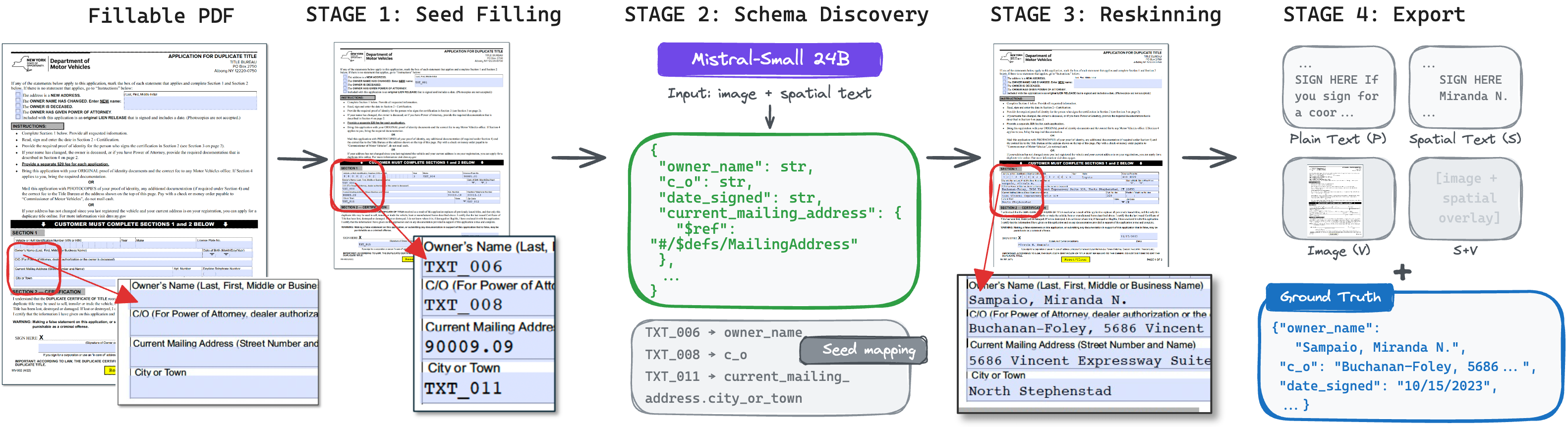}
  \caption{The Reverse Annotation pipeline. \textbf{Stage~1:} Fillable PDF templates are filled with deterministic placeholders (\texttt{TXT\_001}, \texttt{TXT\_002}, \ldots). \textbf{Stage~2:} An LLM discovers a semantic schema by mapping placeholders to field names. \textbf{Stage~3:} Realistic synthetic values replace placeholders and are injected into form widgets. \textbf{Stage~4:} Each filled document is exported in four modalities.}
  \label{fig:pipeline}
\end{figure*}

Structured data extraction from documents (the task of converting forms, invoices, and other structured documents into machine-readable records) is a critical capability for enterprise automation. While multimodal foundation models have shown remarkable progress on document understanding benchmarks, existing evaluation resources have significant limitations. Moreover, many real-world extraction tasks involve straightforward forms where frontier API costs are prohibitive and latency requirements demand on-device inference. Understanding where small models ($\leq$4B) fail (and whether those failures reflect fundamental capability gaps or addressable instruction-following deficits) is critical for guiding efficient model development, yet no existing benchmark systematically evaluates models below 4B parameters on this task.

Current benchmarks such as FUNSD~\cite{jaume2019funsd}, CORD~\cite{park2019cord}, and SROIE~\cite{huang2019sroie} evaluate models on a small number of fixed templates with manually annotated ground truth. More recent efforts like VRDU~\cite{wang2023vrdu} and DocILE~\cite{simsa2023docile} have expanded template diversity but apply a fixed extraction schema across all documents---VRDU uses two schemas and DocILE uses one, limiting evaluation of a model's ability to generalize to unseen document structures. The fundamental challenge is twofold: ensuring annotation accuracy at scale, and evaluating schema variability rather than memorization.

We address this tension with \bench, a benchmark built on a \emph{Reverse Annotation} principle. Instead of annotating existing documents, we start with fillable PDF templates, fill them with deterministic placeholders, use an LLM to discover a semantic schema mapping placeholders to field names, and then inject realistic synthetic values into the form widgets. Because every field value is programmatically written to a known widget (not read from an image), the value-level ground truth is deterministic. The schema-to-widget mapping, however, is LLM-generated and thus fallible: the LLM may misattribute a value to the wrong semantic field (\eg, swapping city and state). Unlike manual annotation errors, these mapping errors are auditable and correctable through the placeholder trace. Through a three-phase QA process combining automated checks, frontier-model audit, and expert review (\cref{sec:qa}), we achieve an estimated ${\sim}$98.5\% field-level accuracy. The rendering pipeline then produces four controlled input representations of each document, enabling systematic study of how input modality affects extraction performance.

\noindent\textbf{Contributions.}
\begin{enumerate}
  \item We present Reverse Annotation, a pipeline for generating synthetic document extraction benchmarks with deterministic value-level ground truth, auditable schema mappings, and controllable modality ablation.
  \item We release \bench, a benchmark of 1,777 documents with 1,771 unique schemas across three structural categories, evaluated on 21,084 fields validated through three-phase quality assurance.
  \item We evaluate 20 models spanning frontier APIs to 800M-parameter open models across four modalities, revealing a taxonomy of small-model failure modes---including schema echo, under-extraction, and an instruction-following threshold between 2--4B---alongside modality and resolution robustness findings.
  \item We define a standardized evaluation protocol with structure-aware reporting (\benchflat, \benchnested, \benchtable, defined by schema structure: no nesting, nested objects, or array-of-objects respectively) to enable meaningful comparison as model capabilities improve.
\end{enumerate}

\section{Related Work}
\label{sec:related}

\vspace{2pt}\noindent\textbf{Document Understanding Benchmarks.}
Early benchmarks used fixed schemas: FUNSD~\cite{jaume2019funsd} (199 forms), CORD~\cite{park2019cord} (11K receipts), and SROIE~\cite{huang2019sroie} (receipts), limiting evaluation of adaptability to new document types. VRDU~\cite{wang2023vrdu} introduced template-diverse extraction across registration and ad-buy forms with two fixed schemas. DocILE~\cite{simsa2023docile} provided 6,680 annotated business documents across 1,152 layout clusters. Both achieve template diversity but rely on manual annotation with unquantified residual error rates, and neither varies the extraction schema across documents.

Concurrent work has moved toward synthetic and programmatic evaluation. SO-Bench~\cite{wang2025sobench} evaluates schema-grounded structured output across four visual domains. ExtractBench~\cite{ferguson2026extractbench} benchmarks end-to-end PDF-to-JSON extraction. JSONSchemaBench~\cite{geng2025jsonschemabench} evaluates constrained decoding on 10K JSON schemas. OmniDocBench~\cite{ouyang2025omnidocbench} provides comprehensive annotations across 9 document types.

Unlike SO-Bench and ExtractBench, which evaluate from a single input representation, \bench combines per-document variable schemas, four controlled modalities, and deterministic ground truth across 1,777 documents (\cref{tab:comparison}).

\begin{table}[t]
  \centering
  \caption{Comparison with publicly available benchmarks. MM\,= multi-modal inputs; SJ\,= structured JSON output. To our knowledge, \bench is the first benchmark combining unique per-document schemas, multi-modal inputs, and structured JSON output.}
  \label{tab:comparison}
  \footnotesize
  \setlength{\tabcolsep}{2.5pt}
  \begin{tabular}{@{}lrcccll@{}}
    \toprule
    Benchmark & Docs & Schemas & MM & SJ & GT Method & Domain \\
    \midrule
    FUNSD & 199 & 1 & \texttimes & \texttimes & Manual & Forms \\
    CORD & 11K & 1 & \texttimes & Part. & Manual & Receipts \\
    VRDU & 3K & 2 & \texttimes & \checkmark & Manual & Gov.\ forms \\
    DocILE & 6.7K & 1 & \texttimes & \texttimes & Hybrid & Business \\
    ExtractB. & 35 & 5 & \texttimes & \checkmark & Manual & Enterprise \\
    \midrule
    \textbf{\bench} & \textbf{1,777} & \textbf{1,771} & \checkmark & \checkmark & \textbf{Rev.\ An.} & \textbf{Gov.\ forms} \\
    \bottomrule
  \end{tabular}
\end{table}

\vspace{2pt}\noindent\textbf{Synthetic Data and Layout Representations.}
The SynthDoG pipeline~\cite{kim2022donut} demonstrates synthetic data's effectiveness for document pre-training; \bench inverts this by filling authentic templates with generated values. Our Spatial Text (plain text with whitespace to preserve column alignment) relates to layout serialization~\cite{luo2025laytextllm,unstract2024llmwhisperer} and the LayoutLM family~\cite{huang2022layoutlmv3,wang2022lilt,xu2020layoutlm,xu2021layoutlmv2}, without special markup.

\section{The VAREX Benchmark}
\label{sec:benchmark}

\subsection{Reverse Annotation Pipeline}
\label{sec:pipeline}

\bench is constructed through a four-stage Reverse Annotation pipeline (\cref{fig:pipeline}) that generates documents from structured data rather than annotating existing ones. The key insight is to decouple value-level ground truth (which is deterministic) from schema-level mapping (which requires LLM inference and validation).

\vspace{2pt}\noindent\textbf{Stage 1: Template Collection and Seed Filling.}
We collect 3,300 fillable PDF form templates from U.S.\ government sources, extracting the first page of each, spanning various federal and state agencies. Of these, 1,946 English-language forms are successfully processed through the full pipeline; the remainder are excluded due to non-English content, multi-page layouts, unfillable widgets, or generation failures at various stages. Forms range from simple 3-field applications to complex documents with nested sections and tabular regions.

Each template's form fields are analyzed using PyMuPDF to extract widget metadata (field types, bounding boxes, fonts, and array grouping patterns). Fields are filled with \emph{deterministic placeholder values} that uniquely identify each widget: text fields receive sequential identifiers (\texttt{TXT\_001}, \texttt{TXT\_002}, \ldots), date fields receive sequential dates from 2099, and numeric fields receive sequential values. These placeholders serve as traceable markers: when a downstream LLM reads the filled form, any placeholder it reports can be traced back to the exact widget that contains it.

\vspace{2pt}\noindent\textbf{Stage 2: Schema Discovery.}
The seed-filled PDF is rendered as an image and its spatial text is extracted; both are passed to a 24B instruction-tuned model\footnote{Mistral-Small-Instruct-2506 (24B); excluded from benchmark evaluation to avoid circular dependency. The evaluated Ministral (14B, \cref{sec:results}) is a distinct model.}, which is prompted to analyze the form's visual layout and field labels and generate a structured JSON Schema. The prompt instructs the model to: (a)~assign semantic field names based on visible labels and context; (b)~group related fields into nested objects (\eg, an Address object with street, city, state, zip); and (c)~detect repeated or tabular sections and represent them as arrays of objects.

After schema discovery, each extracted placeholder is matched back to its source PDF widget via the seed mapping, for example, if the LLM places \texttt{TXT\_042} under the schema path \texttt{applicant\_name}, this creates a traceable link from the semantic field to the specific PDF widget. Post-processing enforces one-to-one field-to-widget mappings, removes boolean fields (checkbox rendering is unreliable), and threads physical constraints (maximum visual characters, choice lists) from the widget metadata into the schema.

This stage is where most ground truth errors originate. The LLM may misattribute a placeholder to the wrong semantic field, fail to detect array structure, or hallucinate fields. We address these failure modes through the three-phase QA process in \cref{sec:qa}.

\vspace{2pt}\noindent\textbf{Stage 3: Data Reskinning.}
Given a schema with traceable field-to-widget mappings, we replace placeholders with realistic synthetic values. Value generation uses two components: (1)~\emph{Persona-based generation} using Python's \texttt{Faker} library with weighted multi-locale distributions to generate diverse pools of names, addresses, phone numbers, and identification numbers. (2)~\emph{LLM-assisted generation} for domain-specific content (\eg, compliance narratives), constrained to the field's schema type and maximum visual character count estimated from the bounding box and font size.

Values are programmatically written to specific PDF widget IDs via PyMuPDF, with post-fill verification to detect write failures and visual truncation (details in supplementary).

\vspace{2pt}\noindent\textbf{Stage 4: Multi-Modal Export.}
Each filled document is exported in four modalities:
\begin{itemize}
  \item \textbf{Plain Text (\mP):} Raw text in reading order via PyMuPDF's \texttt{get\_text()}, with no spatial information.
  \item \textbf{Spatial Text (\mS):} Layout-preserving serialization using whitespace characters to maintain column alignment and field grouping. This approximates the output of layout-aware parsers (\eg, Docling~\cite{livathinos2025docling}).
  \item \textbf{Image (\mV, for Vision):} PNG rendered at 200\,DPI (and 50\,DPI for robustness evaluation).
  \item \textbf{Spatial Text + Image (\mSV):} Both channels provided simultaneously.
\end{itemize}
Because all representations derive from the same filled PDF, any performance difference between modalities reflects the model's processing ability, not information asymmetry. We also release the filled PDFs so researchers can apply their own parsing pipelines.

\subsection{Dataset Composition}
\label{sec:composition}

The final \bench benchmark comprises 299 Flat (17\%), 1,146 Nested (64\%), and 332 Table (19\%) documents, with a median of 11 fields per document and 21,084 total evaluation fields (net of 610 field-level exclusions; see \cref{sec:qa}) spanning 7,042 unique field names, 77\% of which appear in only a single schema, across 1,771 unique schemas (six document pairs share a schema after field normalization; each retains distinct synthetic values and is evaluated independently). Classification is deterministic: a document is \emph{Table} if its schema contains \texttt{"type": "array"} with object items, \emph{Nested} if it contains nested objects but no arrays, and \emph{Flat} otherwise. In practice, the Table category is heterogeneous: 46\% contain multi-column tables ($\geq$2 rows $\times$ $\geq$2 columns), 27\% are single-property lists, and 27\% are single-element arrays; the median has 3 rows.

\begin{figure*}[t]
\centering
\includegraphics[width=\textwidth]{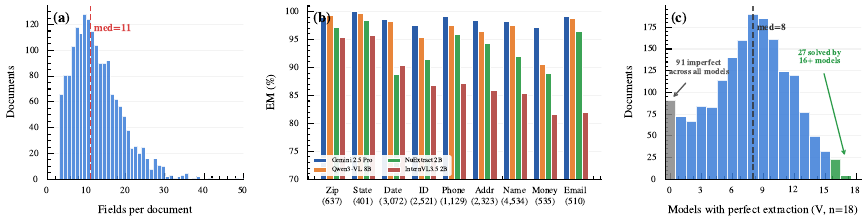}
\caption{Dataset and evaluation overview. (a)~Distribution of extraction fields per document (median 11). (b)~Field-level EM\% by semantic category for four representative models on Image~(\mV); fields are grouped into nine categories by keyword matching (\eg, Name includes \texttt{applicant\_name}, \texttt{witness\_name}, etc.), covering 74\% of 21,084 fields; email and monetary values show the widest cross-scale gaps (15--17\,pp). (c)~Number of vision models (out of 18) achieving perfect extraction per document; 91 documents (5\%) receive imperfect scores from all models, largely attributable to residual annotation noise (see \cref{sec:main_results}).}
\label{fig:dataset}
\end{figure*}

\subsection{Quality Assurance}
\label{sec:qa}

The Reverse Annotation pipeline produces value-level ground truth that is deterministic, every value was programmatically written to a specific widget. However, the \emph{schema mapping} from Stage~2 is LLM-generated and can introduce errors. We conducted a three-phase quality assurance process.

\vspace{2pt}\noindent\textbf{Phase~1 (automated screening).}
Every ground truth value was searched in the extracted text representations to verify it appears in the rendered output. Of 1,946 packaged documents, 1,919 passed (98.6\%); 27 were removed due to value truncation~(18), unreplaced placeholders in the output~(6), or empty ground-truth records~(3). Automated checks also flagged 142 field-level exclusions: 114 schema--ground-truth type mismatches and 28 fields under ambiguous array schemas with empty \texttt{items} definitions.

\vspace{2pt}\noindent\textbf{Phase~2 (frontier-model audit).}
All remaining documents were audited using Claude Sonnet~4.6 as an independent verifier. Of all fields, 96.8\% passed, 2.8\% were flagged as ambiguous, and 0.4\% were automatically excluded as clear errors (included in the Phase~3 exclusion counts below); 428 documents contained at least one ambiguous flag requiring human review.

\vspace{2pt}\noindent\textbf{Phase~3 (expert human review).}
The authors reviewed all 428 flagged documents, resulting in 37 documents removed (systemic generation failures) and 468 fields excluded across 287 documents (rendering artifacts, misattributions, formatting collisions). Field exclusions are applied at scoring time; original GT files are preserved. An additional 60 randomly sampled unflagged documents (${\sim}$660 fields) were reviewed, finding 6 undetected errors, a false negative rate below 1\%.

Combined with the 468 human-review exclusions from Phase~3, the benchmark applies 610 field-level exclusions (2.8\% of all ground-truth fields). A sample-based audit of the top model's remaining errors attributes approximately 0.5--0.7\% of scored fields to residual annotation issues; combined with edge cases not captured by field exclusions, we estimate overall field-level accuracy at ${\sim}$98.5\%. An additional 105 documents were excluded (insufficient scorable fields, non-English content, or systematic generation failures), yielding the final 1,777-document benchmark.

\subsection{Evaluation Protocol}
\label{sec:protocol}

\textbf{Prompt.} Models receive a minimal zero-shot prompt instructing them to extract structured data matching the provided schema and return valid JSON, with \texttt{null} for missing fields (full prompt in supplementary), using \texttt{response\_format: \{"type": "json\_object"\}} and \texttt{temperature=0}. Two exceptions use their recommended prompt formats: h2oVL Mississippi uses the template-based prompt from its model card, and NuExtract~2.0 uses its published extraction format. No prompt tuning or optimization was performed for any model; zero-shot evaluation isolates baseline instruction-following capability, though few-shot prompting may mitigate schema echo in small models (\cref{sec:echo}).

\paragraph{Metrics.} We report exact match (EM) as the primary metric: a field scores 1 if the normalized prediction exactly matches the normalized ground truth, 0 otherwise. We additionally report ANLS (Average Normalized Levenshtein Similarity), which assigns partial credit for near-matches, to distinguish complete misses from minor formatting differences. For array fields, we apply order-invariant matching via the Hungarian algorithm: predicted array elements are optimally assigned to ground-truth elements by maximum field overlap rather than positional index, ensuring models are not penalized for reading table rows in a different traversal order. We designate \mV as the primary evaluation modality and \mP as diagnostic ``hard mode.''


\section{Results and Analysis}
\label{sec:results}

We evaluate 20 models spanning frontier APIs, large open vision-language models (VLMs), small VLMs (2--4B), and low-capacity models ($\leq$2B) (\cref{tab:main}). Of these, 18 support vision input and 16 are evaluated across all four modalities; 2 text-only baselines (Qwen~2.5 72B, Llama~3.3 70B) are evaluated on \mP and \mS only. Open models are served via vLLM~\cite{kwon2023vllm}. h2oVL Mississippi models are evaluated on Image~(\mV) only.

\subsection{Main Results}
\label{sec:main_results}

\Cref{tab:main} presents the primary benchmark results on the Image~(\mV) modality. The benchmark spans an 88\,pp range (9.7\% to 98.0\%), with the strongest model discrimination in the 60--95\% band where architecture, scale, and training choices have the largest impact on accuracy.

\begin{table}[t]
\centering
\caption{\textsc{Varex} Image (V) results. All = EM\% (ANLS\%). Perf.\ = perfect-document rate. \textbf{Bold} = best per column. $^\dagger$Fine-tuned for extraction (base: 9.7\%). $^\ddagger$Schema echo dominates (Sec.~4.4).}
\label{tab:main}
\footnotesize
\setlength{\tabcolsep}{2pt}
\begin{tabular}{@{}llccccc@{}}
\toprule
Model & Size & All & Flat & Nest. & Table & Perf. \\
\midrule
\multicolumn{7}{l}{\textit{Frontier APIs}} \\
Gemini 2.5 Pro~\cite{google2025gemini} & API & \textbf{98.0}\,{\scriptsize(98.5)} & \textbf{97.3} & \textbf{98.2} & \textbf{97.7} & \textbf{82.8} \\
Gemini 2.5 Flash~\cite{google2025gemini} & API & 97.3\,{\scriptsize(97.9)} & 96.9 & 97.7 & 96.5 & 81.0 \\
GPT-4o~\cite{hurst2024gpt4o} & API & 94.8\,{\scriptsize(97.1)} & 95.1 & 95.3 & 93.5 & 64.3 \\
\midrule
\multicolumn{7}{l}{\textit{Large Open ($\geq$8B)}} \\
Qwen3-VL~\cite{bai2025qwen3vl} & 8B & 96.6\,{\scriptsize(97.6)} & 96.6 & 97.3 & 95.0 & 78.2 \\
Llama 4 Mav.~\cite{meta2025llama4} & 17B$\times$128E & 95.6\,{\scriptsize(97.1)} & 95.1 & 96.1 & 94.4 & 68.7 \\
Ministral~\cite{mistral2025} & 14B & 94.8\,{\scriptsize(96.5)} & 94.8 & 95.1 & 94.0 & 66.0 \\
Llama 4 Scout~\cite{meta2025llama4} & 17B$\times$16E & 94.3\,{\scriptsize(95.8)} & 95.2 & 94.3 & 94.1 & 63.4 \\
\midrule
\multicolumn{7}{l}{\textit{Small VLMs (2--4B)}} \\
NuExtract 2.0$^\dagger$~\cite{numind2024nuextract} & 2B & 90.8\,{\scriptsize(92.1)} & 93.1 & 91.9 & 87.3 & 52.7 \\
InternVL3.5~\cite{zhu2025internvl35} & 2B & 85.6\,{\scriptsize(89.1)} & 89.6 & 86.1 & 82.7 & 42.4 \\
Qwen 2.5-VL~\cite{yang2024qwen2vl} & 3B & 82.5\,{\scriptsize(85.4)} & 79.9 & 83.1 & 81.9 & 43.7 \\
Gemma 3n~\cite{gemma2025gemma3} & E2B & 71.0\,{\scriptsize(78.1)} & 76.7 & 71.0 & 68.8 & 19.0 \\
MiniCPM-V4~\cite{yao2024minicpm} & 4B & 67.9\,{\scriptsize(69.8)} & 84.8 & 66.2 & 65.5 & 23.6 \\
Gemma 3~\cite{gemma2025gemma3} & 4B & 65.3\,{\scriptsize(75.0)} & 71.7 & 65.0 & 63.6 & 13.7 \\
h2oVL Miss.~\cite{h2ovl} & 2B & 61.3\,{\scriptsize(64.9)} & 67.2 & 61.3 & 58.9 & 11.6 \\
\midrule
\multicolumn{7}{l}{\textit{Low-capacity ($\leq$2B)$^{\ddagger}$}} \\
Qwen3-VL~\cite{bai2025qwen3vl} & 2B & 34.2\,{\scriptsize(34.7)} & 93.3 & 26.8 & 29.0 & 29.3 \\
h2oVL Miss.~\cite{h2ovl} & 800M & 34.2\,{\scriptsize(36.1)} & 35.8 & 33.2 & 36.0 & 5.9 \\
InternVL3.5~\cite{zhu2025internvl35} & 1B & 28.2\,{\scriptsize(29.8)} & 79.0 & 22.4 & 22.2 & 11.2 \\
Qwen2-VL~\cite{yang2024qwen2vl} & 2B & 9.7\,{\scriptsize(10.0)} & 16.4 & 9.0 & 8.9 & 4.3 \\
\bottomrule
\end{tabular}
\end{table}

Even among models above 90\%, the \benchtable split shows the widest spread: Gemini~2.5~Pro achieves 97.7\% on Table documents while Qwen3-VL~8B drops to 95.0\%, a gap largely masked by aggregate scores. The perfect-document rate further differentiates models: the best model solves 82.8\% of documents perfectly. 91~documents (5\%) receive imperfect scores from all 18 vision models. Manual audit of a sample attributes the majority of top-model errors on these documents to residual annotation issues, consistent with the estimated ${\sim}$1.5\% per-field residual rate. Half of all documents are solved perfectly by 8 or more models (\cref{fig:dataset}c).

Notably, Qwen3-VL 8B (96.6\%) outperforms the much larger Llama~4 Maverick (95.6\%, 17B$\times$128E) and GPT-4o (94.8\%), suggesting that model scale alone does not determine extraction capability. Pairwise differences in this cluster should be interpreted cautiously; at accuracy levels above 95\%, differences of 1\,pp or less may reflect residual ground-truth noise rather than true performance gaps.

Bootstrap confidence intervals (95\%, document-level resampling) confirm that most top-seven pairwise differences are significant ($p < 0.01$), though Ministral 14B, GPT-4o, and Llama~4 Scout form a statistically indistinguishable cluster at 94.3--94.8\%.

ANLS scores (\cref{tab:main}) confirm that most errors are complete misses rather than near-matches: the average EM--ANLS gap across all models is 2.4\,pp, with models exhibiting high non-compliance rates showing near-zero gaps (\eg, Qwen3-VL 2B: 0.5\,pp), while models with genuine extraction errors show larger gaps (\eg, Gemma~3 4B: 9.7\,pp from formatting differences).

\subsection{Structure-Aware Difficulty}
\label{sec:structure}

With order-invariant array matching, document type has little effect on top models: a model scoring above 90\% overall typically scores within 1\,pp on Flat, Nested, and Table documents alike, partly because 54\% of Table documents are trivially extractable single-property lists or single-element arrays (\cref{sec:composition}). The gap between Flat and Table accuracy widens at lower scales: up to 7\,pp for mid-range models (80--90\%) and 8--20\,pp for non-echo models below 80\%, where Table documents expose genuine structural comprehension failures. Even among top models, Table accuracy shows a wider spread across models (93.5\%--97.7\%, 4.2\,pp) than Flat (2.5\,pp), making it the most sensitive split for distinguishing models in the 70--90\% range.

Accuracy varies substantially by semantic category (\cref{fig:dataset}b). Format-sensitive types show the widest gaps: monetary values drop from 97\% (Gemini~2.5 Pro) to 82\% (InternVL3.5 2B), and email addresses from 99\% to 82\%, reflecting precision requirements in decimal formatting and character-level recognition respectively. Simpler types like zip codes and state abbreviations show narrower cross-scale gaps ($<$10\,pp). Among Table errors for large models, the majority involve missing fields or value mismatches rather than structural errors, as the order-invariant array matching ensures models are not penalized for different row traversal orders.

\subsection{Modality Analysis}
\label{sec:modality}

\begin{table}[t]
  \caption{Modality comparison for selected models (EM\%). Best modality per model in \textbf{bold}. Text-only baselines in \emph{italics}.}
  \label{tab:modality}
  \centering
  \small
  \setlength{\tabcolsep}{3pt}
  \begin{tabular}{@{}llcccc@{}}
    \toprule
    Model & Size & \mP & \mS & \mV & \mSV \\
    \midrule
    Gem.\,2.5 Flash\cite{google2025gemini} & API & 93.3 & 96.5 & 97.3 & \textbf{97.8} \\
    GPT-4o\cite{hurst2024gpt4o} & API & 93.0 & 95.8 & 94.8 & \textbf{96.9} \\
    Llama 4 Mav.\cite{meta2025llama4} & 17B$\times$128E & 89.4 & 95.5 & 95.6 & \textbf{97.3} \\
    Qwen3-VL\cite{bai2025qwen3vl} & 8B & 88.0 & 94.4 & 96.6 & \textbf{97.1} \\
    Ministral\cite{mistral2025} & 14B & 85.0 & 92.9 & 94.8 & \textbf{95.5} \\
    InternVL3.5\cite{zhu2025internvl35} & 2B & 75.8 & \textbf{88.9} & 85.6 & 89.4 \\
    Qwen3-VL & 2B & 66.1 & \textbf{73.8} & 34.4 & 33.2 \\
    InternVL3.5 & 1B & 53.3 & \textbf{71.3} & 28.1 & 49.8 \\
    \midrule
    \emph{Qwen 2.5}\cite{yang2024qwen2vl} & \emph{72B} & \emph{91.8} & \emph{\textbf{95.9}} & --- & --- \\
    \emph{Llama 3.3}\cite{meta2025llama4} & \emph{70B} & \emph{89.3} & \emph{\textbf{94.5}} & --- & --- \\
    \bottomrule
  \end{tabular}
\end{table}

\textbf{Choosing an input representation.} The largest accuracy gain comes from upgrading raw text to layout-preserving text. Across all models (\cref{tab:modality}), the \mP$\,\to\,$\mS gain ranges from $+$3 to $+$8\,pp above 90\% EM and up to $+$18\,pp at smaller scales, more than any other single modality change. Naive reading-order extraction discards column alignment and field grouping, and may serialize table columns vertically; whitespace-preserving serialization removes this burden. Once spatial text is available, adding vision yields diminishing returns: \mS$\to$\mV is $-$1.0 to $+$2.1\,pp, and \mV$\to$\mSV is $+$0.5 to $+$2.2\,pp. Text-only models with Spatial Text can match VLMs: Qwen~2.5 72B reaches 95.9\% on \mS, exceeding GPT-4o on Image (94.8\%). Providing both channels (\mSV) is consistently the best or tied-best option. Our Spatial Text derives from digitally native PDFs with precise widget metadata, but a high-quality OCR engine with word-level bounding boxes can produce comparable layout-preserving text from scanned documents.

\subsection{Output Compliance in Small Models}
\label{sec:echo}

\begin{figure}[t]
\centering
\includegraphics[width=\columnwidth]{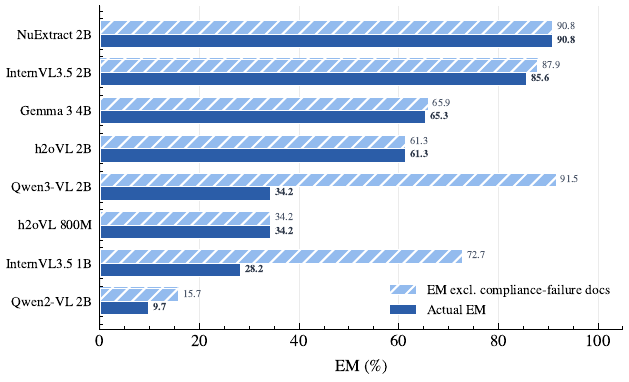}
\caption{Output compliance failures in small models (Image V). Dark bars: actual EM; hatched bars: EM on compliant documents. Failures include schema reproduction (dominant in InternVL3.5 1B) and schema-wrapped extraction (dominant in Qwen3-VL 2B). Qwen3-VL 2B drops from 91.5\% to 34.2\%; InternVL3.5 1B from 72.7\% to 28.2\%. NuExtract and h2oVL show no gap.}
\label{fig:echo}
\end{figure}

Below 4B parameters, extraction accuracy is dominated by \emph{structured output compliance} failures rather than vision or reading errors. \Cref{fig:echo} reports compliance failure rates and conditional accuracy for sub-4B models. We identify two failure modes and a critical scaling threshold.

\vspace{2pt}\noindent\textbf{Schema echo.} Models produce non-compliant JSON that mirrors the input schema rather than containing extracted values. This failure is triggered by JSON Schema's \texttt{\$defs} keyword (reusable type definitions referenced via \texttt{\$ref}): across affected models, non-compliance rates rise from 2--3\% on flat schemas to 68--69\% when \texttt{\$defs} are present (\cref{fig:echo}). The failure takes two forms: \emph{schema reproduction} (returning the definition verbatim, dominant in InternVL3.5 1B) and \emph{schema-wrapped extraction} (correct values wrapped in schema metadata, dominant in Qwen3-VL 2B; see Appendix~C). Image inputs exacerbate both: Qwen3-VL 2B produces non-compliant output on 56\% of documents on images vs.\ 17\% on text, while its outputs frequently contain correct extracted values, confirming that the failure is in formatting, not perception. Inlining \texttt{\$defs} references directly into the schema confirms the causal mechanism: Qwen3-VL 2B rises from 27.4\% to 91.8\% (+64\,pp) on the \texttt{\$defs} subset, and InternVL3.5 1B from 22.3\% to 73.9\% (+52\,pp). This workaround is inapplicable when nested definitions are structurally required, but it confirms that \texttt{\$defs} is the trigger. We retain \texttt{\$defs} in the benchmark because it is the default output of widely used schema generation libraries such as Pydantic; the inlining ablation above provides practitioners a ready workaround.

The most striking case is Qwen3-VL 2B (\cref{fig:echo}): its 91.5\% accuracy on compliant documents would place it between Llama~4 Scout and NuExtract 2.0 in the overall ranking, a 2B model approaching large open model performance when its output formatting is correct, with room for improvement via output post-processing.

h2oVL Mississippi, which receives a JSON template with empty values rather than a schema, exhibits zero echo at both scales, consistent with the hypothesis that schema-like structure triggers the failure.

\vspace{2pt}\noindent\textbf{Under-extraction.} Models that understand the task but lack capacity to complete it produce correct JSON structure with most fields left empty. The h2oVL family exhibits this pattern: zero schema echo at both 800M and 2B, but empty-field rates of 52\% and 27\% respectively. These models show a pronounced \emph{attention decay} pattern, with accuracy on the first quartile of schema fields 2.1$\times$ higher than on the last quartile at 800M and 1.4$\times$ at 2B, suggesting that output generation capacity (not document comprehension) is the binding constraint.

\vspace{2pt}\noindent\textbf{Instruction-following threshold.} Between 2--4B parameters, models transition from output-format failures to genuine extraction errors (OCR mistakes, hallucinated values, missed fields). NuExtract 2.0 2B bypasses this threshold entirely through extraction-specific fine-tuning: zero schema echo and 90.8\% EM (evaluated using its published extraction format) despite a 2B parameter budget. This demonstrates that structured output compliance for document extraction is a learnable skill, not an emergent capability requiring scale, a finding with direct implications for practitioners building small, deployable extraction models.

\subsection{Scaling, Fine-Tuning, and Modality Preference}
\label{sec:modality_pref}

Modality preference varies across model families. At 2B, InternVL3.5 prefers Spatial Text (88.9\% S vs.\ 85.6\% V) while Qwen3-VL scores 73.8\% on S but only 34.2\% on V, however, this gap is largely attributable to image-triggered compliance failures (56\% non-compliant on images vs.\ 17\% on text; see \cref{sec:echo}). When echo documents are excluded, Qwen3-VL 2B achieves 92\% on images and 89\% on plain text, indicating that the apparent modality preference is primarily an instruction-following artifact rather than a vision capability gap.

\vspace{2pt}\noindent\textbf{Scaling and Fine-Tuning.}
The InternVL3.5 family reveals a dramatic scaling threshold: performance jumps from 28.2\% at 1B to 85.6\% at 2B (+57\,pp) on Image. Fine-tuning is equally powerful: NuExtract 2.0 2B gains ${\sim}$+81\,pp over its Qwen2-VL base, transforming a near-non-functional model (9.7\%) into a practical one (90.8\%) with zero schema echo, confirming that the 2--4B instruction-following threshold identified in \cref{sec:echo} can be bypassed entirely through targeted training.

\subsection{Resolution Robustness}
\label{sec:resolution}

\begin{table}[t]
  \caption{Resolution robustness on Image \mV (EM\%).}
  \label{tab:resolution}
  \centering
  \small
  \begin{tabular}{@{}lccc@{}}
    \toprule
    Model & 200\,DPI & 50\,DPI & $\Delta$ \\
    \midrule
    Gemini 2.5 Pro & 98.0 & 96.3 & $-$1.7 \\
    Gemini 2.5 Flash & 97.3 & 93.8 & $-$3.5 \\
    GPT-4o & 94.8 & 65.7 & $-$29.1 \\
    Qwen3-VL 8B & 96.6 & 56.9 & $-$39.7 \\
    Llama 4 Maverick & 95.6 & 57.3 & $-$38.3 \\
    Ministral 14B & 94.8 & 55.1 & $-$39.7 \\
    InternVL3.5 2B & 85.6 & 80.9 & $-$4.7 \\
    \bottomrule
  \end{tabular}
\end{table}

To probe robustness to image quality degradation, we re-evaluate models on images rendered at reduced resolution (50\,DPI, one-quarter the linear resolution of the standard 200\,DPI input), providing a controlled stress test of each model's vision encoder. The results (\cref{tab:resolution}) reveal a sharp divide between API and open models. Gemini models degrade gracefully ($-$1.7 to $-$3.5\,pp), while open models in the 8--17B range suffer catastrophic drops of 38--40\,pp, falling from above 93\% to the mid-50s. GPT-4o loses 29\,pp despite API-level scale. InternVL3.5 2B is a notable exception among open models, losing only 4.7\,pp, suggesting its vision encoder is less dependent on high-frequency detail, consistent with its stronger performance on Spatial Text than on Image inputs (\cref{tab:modality}).

\section{Discussion}
\label{sec:discussion}

\bench's complexity ceiling is set by the schema discovery model (Mistral-Small 24B): nesting rarely exceeds two levels, and ${\sim}$1.5\% of scored fields may contain residual annotation errors. Because schemas are LLM-discovered, the benchmark may underrepresent structures difficult for LLMs to articulate as JSON Schema, potentially biasing toward LLM-friendly document structures. All documents are single-page, English-language U.S.\ government forms with typed values rendered from digital PDFs, lacking handwriting, scan artifacts, and multilingual content. Faker-generated values (\eg, 555-prefix phones) do not match real-world distributions, though this is immaterial to extraction accuracy. Table rows are few (median 3), well below enterprise scale. With order-invariant array matching, row ordering ambiguity is eliminated, and the remaining Table errors reflect genuine structural comprehension failures. Order-invariant matching does not penalize within-row field misattribution, a limitation for applications requiring intra-row correspondence. h2oVL models were evaluated on Image only. The benchmark most effectively discriminates models in the 60--95\% range; above 95\%, residual ground-truth noise limits the precision of model-to-model comparisons. \bench evaluates publicly available form templates with synthetically generated values, introducing no privacy concerns. All data and evaluation scripts are released under permissive licenses.

\section{Conclusion}
\label{sec:conclusion}

We presented \bench, a benchmark for structured extraction from document images with 1,777 documents, 1,771 unique schemas, and 21,084 evaluation fields. The Reverse Annotation pipeline produces deterministic value-level ground truth validated through three-phase quality assurance, across four controlled input modalities, enabling systematic study of how models process different document representations. Our analysis of 20 models from 800M to frontier scale identifies concrete failure modes---schema echo, under-extraction, and attention decay---that offer clear targets for improving sub-4B models, a regime critical for cost-sensitive and on-device deployment yet largely unexamined by existing benchmarks. The benchmark spans an 88\,pp performance range and most effectively discriminates models in the 60--95\% band, where small-model failures are dominated by structured output compliance deficits rather than vision limitations. We hope \bench will serve as both a diagnostic tool for understanding extraction capabilities and a catalyst for developing efficient, deployable extraction models.

{
    \small
    \bibliographystyle{ieeenat_fullname}
    \bibliography{main}
}

\onecolumn
\raggedbottom
\appendix

\lstset{basicstyle=\ttfamily\small,breaklines=true,breakatwhitespace=true,aboveskip=0.5em,belowskip=0.5em,xleftmargin=1em,framexleftmargin=0em}

\ifdefined\supplmode\else
\begin{center}
{\Large\bfseries Supplementary Material}
\end{center}
\vspace{4pt}
\fi

\noindent This supplement provides additional detail and examples supporting the main paper:
\begin{itemize}[leftmargin=*,itemsep=2pt]
  \item \textbf{Appendix A} --- Evaluation prompt used for all models.
  \item \textbf{Appendix B} --- Complete document examples: a Nested-category form (B.1) and a Table-category form (B.2), each with schema and ground truth.
  \item \textbf{Appendix C} --- Schema echo examples: three models receiving the same document, illustrating the compliance failure spectrum.
  \item \textbf{Appendix D} --- Resolution robustness: 200\,DPI vs.\ 50\,DPI visual comparison with per-field error analysis.
  \item \textbf{Appendix E} --- Input modality comparison: Plain Text vs.\ Spatial Text representations of the same document.
\end{itemize}

\noindent\textbf{A.\quad Evaluation Prompt.}
The full evaluation prompt used for all models (except h2oVL Mississippi and NuExtract~2.0, which use their published formats):

\begin{lstlisting}
System: Extract structured data from this document. Return a JSON object matching this schema: {schema}
Return null for fields you cannot find.
Return ONLY valid JSON.
Return an instance of the JSON with extracted values, not the schema itself.
\end{lstlisting}

\noindent\textbf{B.\quad Document Examples.}
We present two complete examples: a Nested document illustrating \texttt{\$defs}/\texttt{\$ref} schema structure, and a Table document illustrating array-of-objects extraction.

\vspace{4pt}
\noindent\textbf{B.1.\quad Nested Example: Housing Terms and Conditions.}
A Nested-category document illustrating \texttt{\$defs}/\texttt{\$ref} schema structure.

\begin{center}
\fbox{\includegraphics[width=0.79\textwidth]{housing_terms_nested.png}}
\captionof{figure}{Example \bench document (Nested category). The schema uses \texttt{\$defs}/\texttt{\$ref} to define reusable nested object types (HousingProvider, IndividualInCharge, HousingFacilityMailingAddress). Only the English-language fields contain fillable widgets; the Spanish translation serves as static context.}
\label{app:nested_example}
\end{center}

\clearpage
\noindent\textbf{Schema:}
\begin{lstlisting}[language=json]
{
  "type": "object",
  "$defs": {
    "HousingFacilityMailingAddress": {
      "type": "object",
      "properties": {
        "address": {"type": "string", "description": "Mailing address of the housing facility"},
        "city_state_zip": {"type": "string", "description": "City, state, and zip code"},
        "phone": {"type": "string", "description": "Phone number"}
      }
    },
    "HousingProvider": {
      "type": "object",
      "properties": {
        "name": {"type": "string", "description": "Name of the housing provider"},
        "address": {"type": "string", "description": "Address of the housing provider"},
        "city_state_zip": {"type": "string", "description": "City, state, and zip code"}
      }
    },
    "IndividualInCharge": {
      "type": "object",
      "properties": {
        "name": {"type": "string", "description": "Name of the individual in charge"},
        "address": {"type": "string", "description": "Address of the individual in charge"},
        "city_state_zip": {"type": "string", "description": "City, state, and zip code"},
        "phone": {"type": "string", "description": "Phone number"}
      }
    }
  },
  "properties": {
    "conditions_of_occupancy": {"type": "string", "description": "Conditions of occupancy"},
    "housing_provider": {"$ref": "#/$defs/HousingProvider"},
    "individual_in_charge": {"$ref": "#/$defs/IndividualInCharge"},
    "housing_facility_mailing_address": {"$ref": "#/$defs/HousingFacilityMailingAddress"}
  }
}
\end{lstlisting}

\noindent\textbf{Ground Truth:}
\begin{lstlisting}[language=json]
{
  "conditions_of_occupancy": "No smoking, no pets allowed",
  "housing_provider": {
    "name": "Knox Inc",
    "address": "41516 Anita Plaza Suite 990",
    "city_state_zip": "Jasmineberg, TX 44064"
  },
  "individual_in_charge": {
    "name": "Joseph, Garima C.",
    "address": "41516 Anita Plaza Suite 990",
    "city_state_zip": "Jasmineberg, TX 44064",
    "phone": "(283) 690-4093"
  },
  "housing_facility_mailing_address": {
    "address": "41516 Anita Plaza Suite 990",
    "city_state_zip": "Jasmineberg, TX 44064",
    "phone": "(337) 134-7229"
  }
}
\end{lstlisting}

\noindent Each value was programmatically written to a specific PDF widget and is deterministic. The nested structure mirrors the \texttt{\$defs} definitions in the schema.

\clearpage

\noindent\textbf{B.2.\quad Table Example: ALJ Application.}
A Table-category document illustrating array-of-objects extraction with value truncation.

\begin{center}
\fbox{\includegraphics[width=0.79\textwidth]{alj_application_table.png}}
\captionof{figure}{Example \bench document (Table category). The schema contains both nested objects (\texttt{Address}) and arrays of objects (\texttt{alj\_experience}, \texttt{licensure\_states}). Ground-truth values are trimmed to match the visible rendered text when PDF widget bounding boxes truncate the content (\eg, \texttt{Departmen} rather than \texttt{Department}), ensuring models are evaluated against what is actually readable.}
\label{app:table_example}
\end{center}

\clearpage

\noindent\textbf{Schema (condensed, field descriptions omitted):}
\begin{lstlisting}[language=json]
{
  "type": "object",
  "$defs": {
    "ALJExperience": {
      "type": "object",
      "properties": {
        "position": {"type": "string"}, "agency": {"type": "string"},
        "start_date": {"type": "string"}, "end_date": {"type": "string"}
      }
    },
    "Address": {"type": "object", "properties": {"street": {"type": "string"}}},
    "LicensureState": {"type": "object", "properties": {"state": {"type": "string"}}}
  },
  "properties": {
    "applicant_name": {"type": "string"}, "last_five_ssn_digits": {"type": "string"},
    "daytime_phone": {"type": "string"}, "other_phone": {"type": "string"},
    "email": {"type": "string"}, "highest_grade_step_level_rate": {"type": "string"},
    "retirement_date": {"type": "string"}, "interested_agencies": {"type": "string"},
    "address": {"$ref": "#/$defs/Address"},
    "alj_experience": {"type": "array", "items": {"$ref": "#/$defs/ALJExperience"}},
    "licensure_states": {"type": "array", "items": {"$ref": "#/$defs/LicensureState"}}
  }
}
\end{lstlisting}

\noindent\textbf{Ground Truth:}
\begin{lstlisting}[language=json]
{
  "applicant_name": "Renee T. Estrada",
  "last_five_ssn_digits": "09801",
  "daytime_phone": "2109652",
  "other_phone": "2790389",
  "email": "restrada@campbell.com",
  "highest_grade_step_level_rate": "Administrative Law Judge, Grade 15, Step 5, $145,600",
  "retirement_date": "06/15/2023",
  "interested_agencies": "Department of Veterans Affairs, Social Security Administration, Department of Labor",
  "address": {"street": "1544 Michael Burgs Suite 985"},
  "alj_experience": [
    {"position": "Administrativ", "agency": "Departmen", "start_date": "01/15/2005", "end_date": "06/15/2023"},
    {"position": "Hearings", "agency": "Social", "start_date": "03/10/2000", "end_date": "01/14/2005"},
    {"position": "Attorney", "agency": "Departmen", "start_date": "07/20/1995", "end_date": "03/09/2000"}
  ],
  "licensure_states": [{"state": "California"}]
}
\end{lstlisting}

\clearpage

\noindent\textbf{C.\quad Schema Echo Examples.}
Three models receive the document in \cref{fig:echo-input} with the schema shown below. We show the first lines of each output to illustrate the spectrum of structured output compliance failures described in Sec.~4.4.

\begin{center}
\fbox{\includegraphics[width=0.79\textwidth]{lab167os.png}}
\captionof{figure}{Input document for the schema echo comparison. A laboratory certification form with both flat fields and array-of-objects structures (\texttt{test\_volumes}, \texttt{other\_tests}) defined via \texttt{\$defs}/\texttt{\$ref}. Three models receive this document with the schema below; their outputs illustrate the compliance failure spectrum described in Sec.~4.4.}
\label{fig:echo-input}
\end{center}

\clearpage
\noindent\textbf{Schema} (condensed, field descriptions omitted):
\begin{lstlisting}[language=json]
{
  "type": "object",
  "$defs": {
    "OtherTest": {"type": "object", "properties": {
        "test_name": {"type": "string"}, "annual_volume": {"type": "string"}}},
    "TestVolume": {"type": "object", "properties": {
        "annual_volume": {"type": "string"}}}
  },
  "properties": {
    "laboratory_name": {"type": "string"}, "state_id_number": {"type": "string"},
    "clia": {"type": "string"}, "address": {"type": "string"},
    "city": {"type": "string"}, "state": {"type": "string"},
    "zip_code": {"type": "string"}, "sub_total_other_tests": {"type": "string"},
    "sub_total_specialties": {"type": "string"}, "total_volume": {"type": "string"},
    "printed_name": {"type": "string"}, "signature_date": {"type": "string"},
    "test_volumes": {"type": "array", "items": {"$ref": "#/$defs/TestVolume"}},
    "other_tests": {"type": "array", "items": {"$ref": "#/$defs/OtherTest"}}
  }
}
\end{lstlisting}

\noindent\textbf{Gemini 2.5 Pro (correct extraction):}
\begin{lstlisting}
{
  "laboratory_name": "Parker Diagnostic Laboratories",
  "state_id_number": "LAB-2023-CO-",
  "clia": "01D2345678",
  "address": "7862 Davis Tunnel",
  "city": "Karenview",
  "state": "NJ",
  ...
}
\end{lstlisting}

\noindent\textbf{Qwen3-VL 2B (envelope echo, values extracted but wrapped in schema metadata):}
\begin{lstlisting}
{
  "type": "object",
  "properties": {
    "laboratory_name": "Parker Diagnostic Laboratories",
    "state_id_number": "LAB-2023-CO-",
    "clia": "01D2345678",
    "address": "7862 Davis Tunnel",
    ...
  }
}
\end{lstlisting}

\noindent\textbf{InternVL3.5 1B (pure schema echo, schema definition reproduced verbatim):}
\begin{lstlisting}
{
  "laboratory_name": {
    "type": "string",
    "description": "Name of the laboratory"
  },
  "state_id_number": {
    "type": "string",
    "description": "State ID number of the laboratory"
  },
  ...
}
\end{lstlisting}

\vspace{4pt}
\noindent Three model outputs on the same document (Image \textbf{V} input). Gemini 2.5 Pro extracts values correctly. Qwen3-VL 2B extracts the correct values but wraps them in the schema's structural metadata (\texttt{"type": "object"}, \texttt{"properties": \{...\}}), producing invalid output. InternVL3.5 1B reproduces the schema definition verbatim with no extracted values. Both failure modes are triggered by \texttt{\$defs}/\texttt{\$ref} in the input schema.

\clearpage

\noindent\textbf{D.\quad Resolution Robustness Example.}
Table~4 reports that open models in the 8--17B range lose 38--40\,pp when image resolution drops from 200\,DPI to 50\,DPI, while Gemini models lose under 3.5\,pp. We illustrate this with a Table-category document (\cref{fig:res-200,fig:res-50}): a bond loss application containing 11 array rows of serial numbers, dates, and denominations. At 50\,DPI, digits and dates become visually ambiguous, producing the OCR-like errors shown in \cref{tab:res-errors}.

\noindent\textbf{Schema} (condensed, field descriptions omitted):
\begin{lstlisting}[language=json]
{
  "type": "object",
  "$defs": {
    "Bond": {
      "type": "object",
      "properties": {
        "bond_identifier": {"type": "string"},
        "denomination": {"type": "number"},
        "issue_date": {"type": "string"},
        "description": {"type": "string"}
      }
    }
  },
  "properties": {
    "customer_name": {"type": "string"},
    "customer_number": {"type": "string"},
    "film_record_unavailable_reason": {"type": "string"},
    "disappearance_circumstances": {"type": "string"},
    "bonds": {"type": "array", "items": {"$ref": "#/$defs/Bond"}}
  }
}
\end{lstlisting}

\begin{figure}[p]
\centering
\fbox{\includegraphics[width=0.92\textwidth]{sav2517_200dpi.png}}
\caption{\bench document at standard resolution (200\,DPI). A Table-category bond loss application with 11 array rows. All top models achieve 100\% EM at this resolution.}
\label{fig:res-200}
\end{figure}

\begin{figure}[p]
\centering
\fbox{\includegraphics[width=0.92\textwidth]{sav2517_50dpi.png}}
\caption{Same document at reduced resolution (50\,DPI). Serial numbers, dates, and small text become visually ambiguous. Open models in the 8--17B range lose 13--35\,pp; Gemini models remain at 100\%.}
\label{fig:res-50}
\end{figure}

\begin{table}[h]
\centering
\caption{Representative extraction errors at 50\,DPI on the document in \cref{fig:res-200,fig:res-50}. All models shown score 100\% at 200\,DPI. Error types at 50\,DPI: digit confusion, character substitution, and missed fields.}
\label{tab:res-errors}
\small
\begin{tabular}{@{}llll@{}}
\toprule
Model & Field & Ground Truth & Prediction \\
\midrule
\multirow{3}{*}{Maverick 17B$\times$128E}
  & bonds[1].issue\_date & 05/22/2012 & 05/20/2012 \\
  & bonds[1].bond\_id & BBB 98765432 B & BBB 8765432 B \\
  & customer\_number & DPM-2020-16514 & DPM 2020-16514 \\
\midrule
\multirow{2}{*}{Qwen3-VL 8B}
  & customer\_name & Nadia I. Jaimes & Nadia J. Jaimes \\
  & customer\_number & DPM-2020-16514 & DPM-2020-18514 \\
\midrule
\multirow{2}{*}{Ministral 14B}
  & customer\_name & Nadia I. Jaimes & Nadia I. James \\
  & bonds[6].issue\_date & 02/28/2021 & 02/08/2021 \\
\midrule
Gemini 2.5 Pro & \multicolumn{3}{c}{\textit{100\% EM at both resolutions}} \\
Gemini 2.5 Flash & \multicolumn{3}{c}{\textit{100\% EM at both resolutions}} \\
\bottomrule
\end{tabular}
\end{table}

\clearpage

\noindent\textbf{E.\quad Input Modality Comparison: Plain Text vs.\ Spatial Text.}
The four \bench modalities provide controlled variation in what information the model receives. The largest accuracy gain comes from adding spatial layout to plain text ($+$3--18\,pp; Sec.~4.3). Below we show both text representations for the Table-category document in \cref{app:table_example} (Appendix~B.2), focusing on the region where the difference is most pronounced.

In practice, Spatial Text can be reconstructed from any OCR engine that provides word-level bounding boxes (including lightweight CPU-based systems) without requiring a vision-language model. This makes layout-preserving text a practical input option even in cost-constrained deployments: a small text-only model receiving Spatial Text can approach the accuracy of a larger VLM receiving the document image (Sec.~4.3, Tab.~3).

\begin{center}
\includegraphics[width=0.82\textwidth]{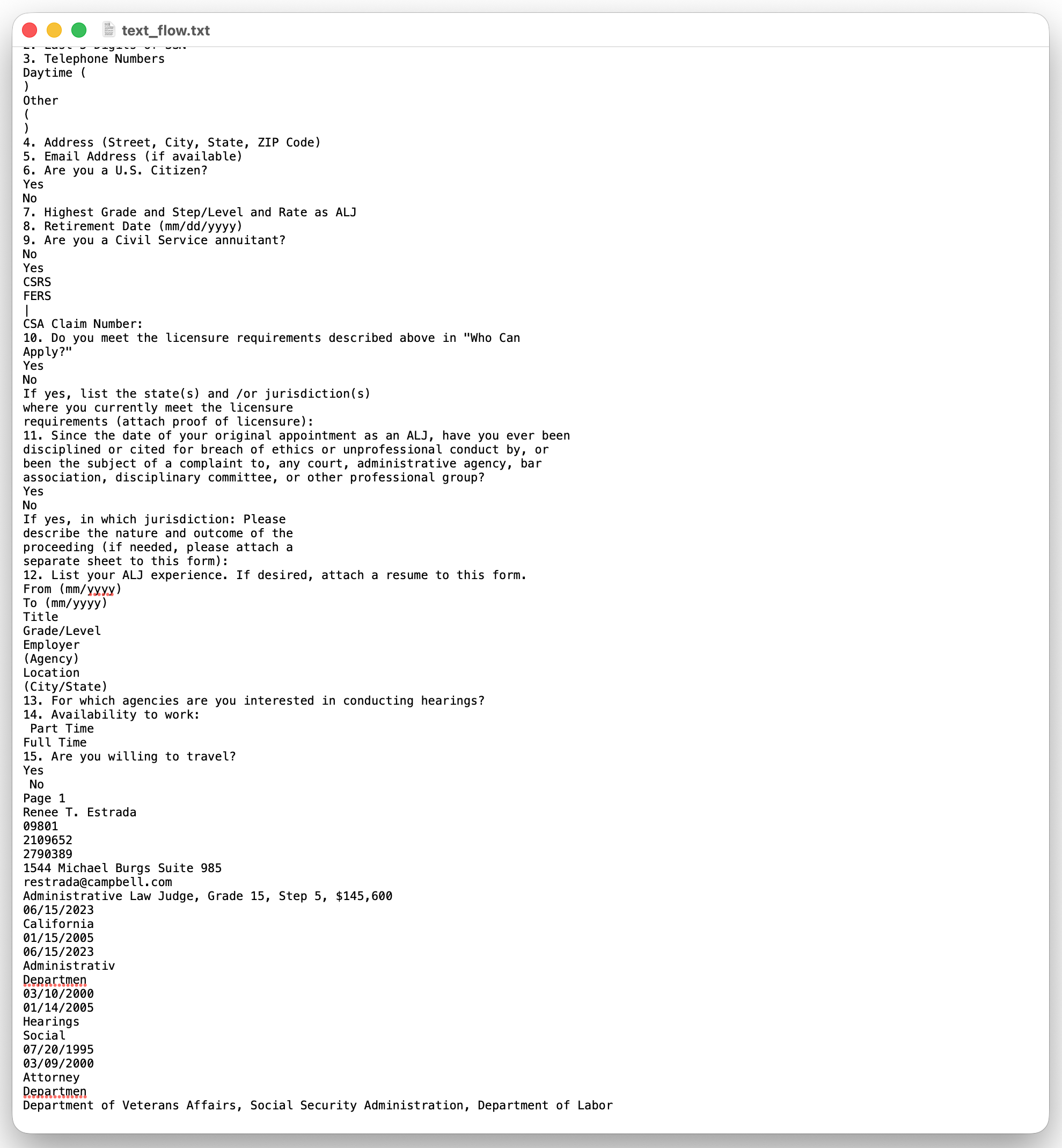}
\captionof{figure}{Plain Text (\textbf{P}) representation of the ALJ form (\cref{app:table_example}), produced by PyMuPDF \texttt{get\_text()} in reading order. The form header is cropped for space. Note that table columns are serialized vertically, all ``From'' dates appear in sequence, then all ``To'' dates, then all titles. Large models can resolve this ordering, but smaller models struggle: the \mP$\,\to\,$\mS gain ranges from $+$3\,pp for frontier models to $+$18\,pp at smaller scales (Tab.~3).}
\label{fig:plain-text}
\end{center}

\begin{center}
\includegraphics[width=0.82\textwidth]{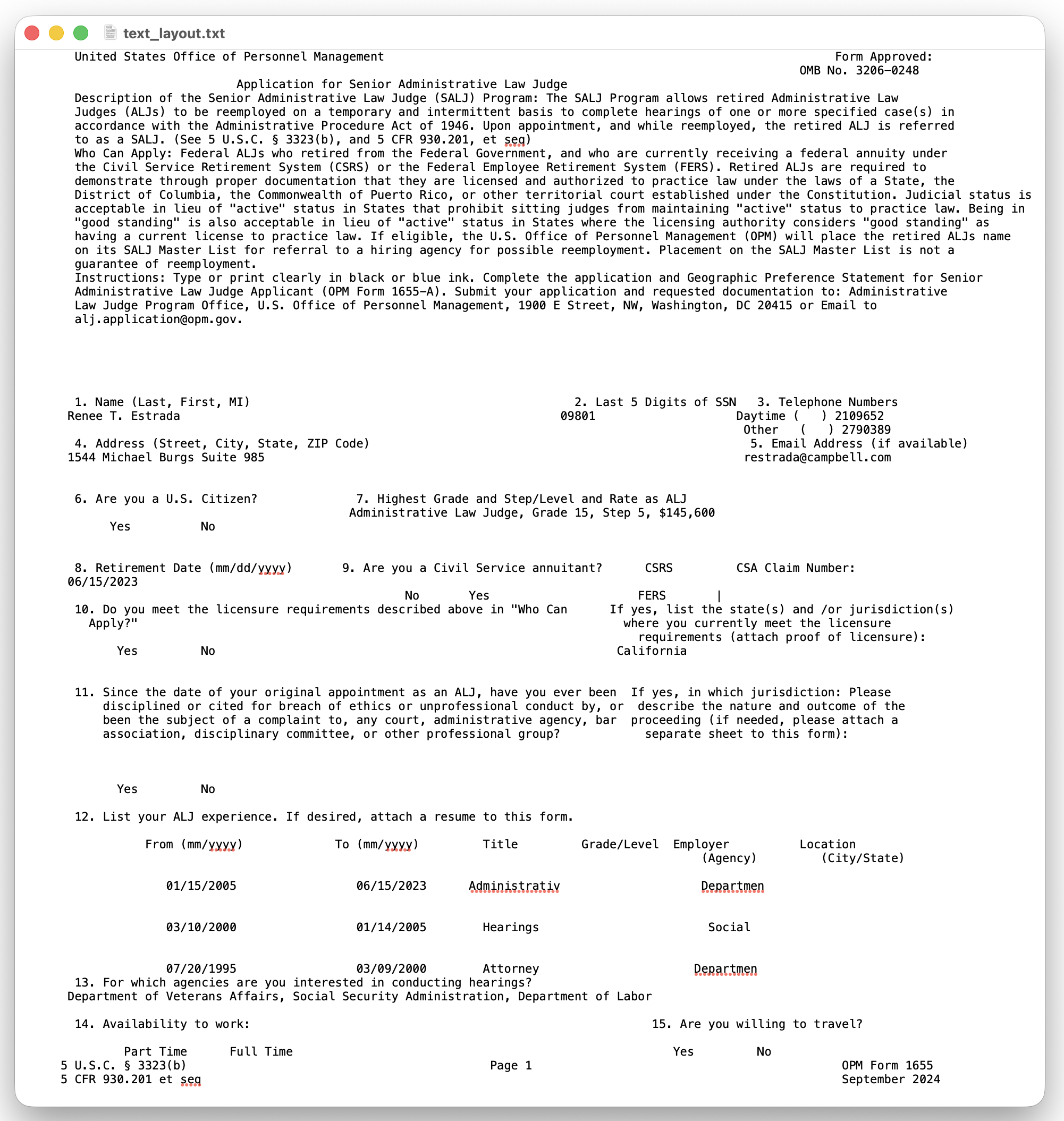}
\captionof{figure}{Spatial Text (\textbf{S}) representation of the same ALJ form. Whitespace characters preserve column alignment, allowing the model to associate values within the same table row (\eg, \texttt{01/15/2005}, \texttt{06/15/2023}, \texttt{Administrativ}, \texttt{Departmen} on the same line). Compare with the Plain Text in \cref{fig:plain-text}, where these values are serialized in separate column runs. This spatial structure accounts for the $+$3--18\,pp accuracy gain reported in Sec.~4.3.}
\label{fig:spatial-text}
\end{center}

\end{document}